\documentclass[11pt,a4paper]{article}
\usepackage[utf8]{inputenc}
\usepackage[hyperref]{acl2020}
\usepackage{times}
\usepackage{latexsym}
\usepackage{booktabs} 
\usepackage{color}
\usepackage{url}
\usepackage{balance}
\usepackage{amsmath}
\usepackage{amsfonts}
\usepackage{graphicx}
\usepackage{amssymb, bbm}
\usepackage{multirow, multicol}
\usepackage{caption, subcaption}
\usepackage{comment}
\usepackage{xspace}
\usepackage{upquote}
\usepackage{float}
\usepackage{alltt}
\usepackage{algpseudocode}
\usepackage{pbox}
\usepackage{enumitem}
\usepackage{mathpartir}
\usepackage{wrapfig}
\usepackage{array}
\usepackage{hhline}
\usepackage{stmaryrd}
\usepackage{pifont}
\usepackage{makecell}
\usepackage{cuted}

\aclfinalcopy 
\def\aclpaperid{2078} 

\definecolor{comment-red}{rgb}{0.8,0,0}

\newcommand{\Red}[1]{{\color{comment-red} #1}}

\newcommand{\Greet}[0]{{\small \textsc{Greet}}\xspace}
\newcommand{\SearchRequest}[0]{{\small \textsc{SearchRequest}}\xspace}
\newcommand{\CompleteRequest}[0]{{\small \textsc{CompleteRequest}}\xspace}
\newcommand{\CompleteTransaction}[0]{{\small \textsc{CompleteTransaction}}\xspace}
\newcommand{\CloseConversation}[0]{{\small \textsc{CloseConversation}}\xspace}
\newcommand{\GreetUser}[0]{{\small \textsc{GreetUser}}\xspace}
\newcommand{\AskQuestion}[0]{{\small \textsc{AskQuestion}}\xspace}
\newcommand{\Answer}[0]{{\small \textsc{Answer}}\xspace}
\newcommand{\OfferReservation}[0]{{\small \textsc{OfferReservation}}\xspace}
\newcommand{\AskByName}[0]{{\small \textsc{AskByName}}\xspace}
\newcommand{\AddConstraints}[0]{{\small \textsc{AddConstraints}}\xspace}
\newcommand{\Accept}[0]{{\small \textsc{Accept}}\xspace}
\newcommand{\Reject}[0]{{\small \textsc{Reject}}\xspace}
\newcommand{\StartState}[0]{{\small \textsc{Start}}\xspace}
\newcommand{\EndState}[0]{{\small \textsc{End}}\xspace}
\newcommand{\SlotQuestion}[0]{{\small \textsc{ SlotQuestion}}\xspace}
\newcommand{\ProposeEntity}[0]{{\small \textsc{ProposeEntity}}\xspace}
\newcommand{\AskSlotQuestion}[0]{{\small \textsc{AskSlotQuestion}}\xspace}

\definecolor{dkgreen}{rgb}{0,0.6,0}
\definecolor{gray}{rgb}{0.5,0.5,0.5}
\definecolor{mauve}{rgb}{0.58,0,0.82}

\title{Zero-Shot Transfer Learning with Synthesized Data \\
for Multi-Domain Dialogue State Tracking
}

\author{Giovanni Campagna \quad Agata Foryciarz \quad Mehrad Moradshahi \quad Monica S. Lam \\
  Computer Science Department \\
  Stanford University \\
  Stanford, CA, USA \\
  \texttt{\{gcampagn,agataf,mehrad,lam\}@cs.stanford.edu} \\}

\date{}

\begin{document}
\maketitle

\begin{abstract}
Zero-shot transfer learning for multi-domain dialogue state tracking can allow us to handle new domains without incurring the high cost of data acquisition. 
This paper proposes new zero-short transfer learning technique for dialogue state tracking where the in-domain training data are all synthesized from an abstract dialogue model and the ontology of the domain.
We show that data augmentation through synthesized data can improve the accuracy of zero-shot learning for both the TRADE model and the BERT-based SUMBT model on the MultiWOZ 2.1 dataset. We show training with only synthesized in-domain data on the SUMBT model can reach about 2/3 of the accuracy obtained with the full training dataset. 
 We improve the zero-shot learning state of the art on average across domains by 21\%.
 
\end{abstract}

\section{Introduction}

\begin{figure}[htb]
\small
\begin{tabbing}
\=123\=\kill
\>S:\>$\textsc{Start}$\\
\>U:\>Can you help with information regarding a food place?\\
\>\>I need to book at 15:45.\\
\>S:\>$\textsc{SearchRequest}\ \text{restaurant}(\text{book\_time}=\text{``15:45''})$\\
\>A:\>How about the restaurant with name La Tasca and \\
\>\>Italian food?\\
\>U:\>Can you find something which serves seafood?\\
\>S:\>$\textsc{SearchRequest}\ \text{restaurant}(\text{book\_time}=\text{``15:45''},$\\ \>\>$\text{food}=\text{``seafood''})$\\
\>A:\>What date are you looking for?\\
\>U:\>Thursday please.\\
\>S:\>$\textsc{SearchRequest}\ \text{restaurant}(\text{book\_time}=\text{``15:45''},$\\ \>\>$\text{food}=\text{``seafood''}, \text{book\_day}=\text{``thursday''})$\\
\>A:\>How about the Copper Kettle? It is a food place with\\
\>\>seafood food.\\
\>U:\>What is the price range and the area?\\
\>S:\>$\textsc{SlotQuestion}\ \text{restaurant}(\text{book\_time}=\text{``15:45''},$\\ \>\>$\text{food}=\text{``seafood''}, \text{book\_day}=\text{``thursday''},$\\
\>\>$\text{price\_range}=?, \text{area}=?)$\\
\>A:\>The Copper Kettle is a moderately priced restaurant in\\
\>\>the north of the city. Would you like a reservation?\\
\>U:\>No, thanks.\\
\>S:\>$\textsc{Close}\ \text{restaurant}(\text{book\_time}=\text{``15:45''},$\\ \>\>$\text{food}=\text{``seafood''}, \text{book\_day}=\text{``thursday''})$\\
\>A:\>Can I help with you anything else?\\
\>U:\>Thank you, that will be it for now.\\
\>S:\>$\textsc{End}\ \text{restaurant}(\text{book\_time}=\text{``15:45''},$\\ \>\>$\text{food}=\text{``seafood''}, \text{book\_day}=\text{``thursday''})$\\
\end{tabbing}
\vspace{-0.6cm}
\caption{An example of a dialogue that can be synthesized from our templates.
`U:' indicates the user, `A:` the agent, and `S:` is the dialogue state at each turn.}
\label{fig:dialogue-example}
\vspace{-0.5cm}
\end{figure}
Automated conversational agents can reduce the costs of customer support, a necessary service in just about every business.  However, training a goal-directed dialogue agent for a domain often requires acquiring annotated dialogues to cover all possible conversation flows. Commonly, this is done using the Wizard-of-Oz technique~\cite{kelley1984iterative}, where two crowdsource workers converse with each other, while also annotating the state at each turn. This technique has been employed to construct several datasets~\cite{hemphill1990atis, wen2016network, yu2019cosql}. Recently, it has been used to build the MultiWOZ dataset~\cite{budzianowski2018large}, a large corpus of dialogues across 7 domains.

Unfortunately, not only is the initial acquisition expensive, annotating dialogues correctly has proven to be challenging due to human errors, delays in annotation, inconsistent conventions, and normalization issues~\cite{eric2019multiwoz, zhou2019multi}.  The MultiWOZ dataset still has significant inconsistencies~\cite{zhou2019multi} despite having been constructed through multiple rounds of annotations~\cite{budzianowski2018large,eric2019multiwoz}. 

We observe empirically from the MultiWOZ training data that conversations in all the domains follow the same pattern: the agent and user start by greeting each other, tehn they converse to find a proposal that satisfies the user, the user provides additional required information, and finally the agent completes the user's transaction.

To facilitate transfer learning, we create an abstract model of dialogues that is independent of the domain of the conversation. In this paper we will focus on dialogues for transactions; other kinds of dialogues such as opinion sharing will have different models.  We have developed an algorithm that accepts an ontology of a domain and a few phrases commonly used in that domain. The algorithm synthesizes dialogue training data based on an abstract dialogue model. 
The dialogue synthesized consists of turns of conversation, each of which has a start state, an agent utterance, a user utterance, and an end state. The start and end states summarize the semantics of the conversation at those points. 
 An example of a dialogue that can be synthesized by our model is shown in Fig.~\ref{fig:dialogue-example}.

To transfer knowledge to a new domain in a zero-shot setting, we train with the synthesized data for the new domain together with existing data for other domains. In addition, we adapt training samples from related domains by substituting them with the vocabulary of the new domain.  We can improve the accuracy of the abstract dialogue model as well as the state-tracking neural network by iteratively refining the model based on the error analysis on the validation data, and by introducing additional annotations in the new domain. 
Note that the abstract dialogue model can be also used directly to implement the agent itself. 

The contributions of this paper are as follows:
\begin{itemize}[topsep=0pt,itemsep=-3pt]
\item A new zero-short transfer learning technique for dialogue state tracking where the in-domain training data are all synthesized from an abstract dialogue model and the ontology of the domain.  


\item Our approach improves over the previous state-of-the-art result on zero-shot transfer learning for MultiWOZ 2.1 tasks by 21\% on average across domains.

\item We show that our approach improves the accuracy for TRADE~\cite{Wu2019May}, an RNN-based model, and SUMBT~\cite{Lee2019Jul}, a BERT-based model~\cite{Devlin2018Oct}, suggesting that our technique is independent of the specific model used.

\item Our experimental results show that synthesized data complements BERT pretraining. The BERT-based SUMBT model can, in a purely zero-shot fashion, achieve between 61\% and 92\% of the accuracy obtained by a model trained on the full dataset. We propose combining pretrained models with synthesized data as a general technique to bootstrap new dialogue state trackers.
\end{itemize}

\section{Related Work}
\label{sec:related}

\paragraph{Dialogue Datasets and Synthesis.}
Synthesized data (in training and evaluation) was proposed by \newcite{weston2015towards} to evaluate the ability of neural models to reason compositionally, and was also used in visual question answering~\cite{Johnson_2017_CVPR, Hudson_2019_CVPR} and semantic parsing~\cite{lake2018generalization}.

\newcite{overnight} proposed synthesizing data, then crowdsourcing paraphrases to train semantic parsers. Various semantic parsing datasets have been generated with this technique~\cite{su2017building, zhong2017seq2sql} and the technique has also been adapted to the multiturn setting~\cite{DBLP:journals/corr/abs-1812-10037, DBLP:journals/corr/abs-1801-04871}. While it tends to be well-annotated, paraphrase data is expensive to acquire, and these datasets are very small.

More recently, we proposed training with both a large amount of synthesized data and a small amount of paraphrase data for semantic parsing of single sentences~\cite{geniepldi19, xu2020schema2qa}. We showed that training with such data can perform well on real-world evaluations. This paper extends this work to the multi-turn setting. Dialogues are more complex as they need to capture information, such as the abstract dialogue state, that is not present in the target annotation (domain and slot values). We extend the synthesis algorithm to operate based on a dialogue model, tracking enough information to continue the dialogue. We also present a novel dialogue model that is suitable for synthesis.

\paragraph{Dialogue State Tracking.}
Dialogue state tracking is a long-studied field, starting with the first \textit{Dialogue State Tracking Challenge}~\cite{williams2014dialog}. A review of prior work can be found by~\newcite{williams2016dialog}.

Previous works on DST use different approaches, ranging from using handcrafted features to elicit utterance information \cite{henderson2014robust, wang2013simple}. \newcite{mrkvsic2016neural} use Convolutional Neural Networks to learn utterance representations. However, their models do not scale as they do not share parameters across different slots. \newcite{zhong2018global} and \newcite{nouri2018toward} propose a new global module that shares information to facilitate knowledge transfer. However, they rely on a predefined ontology. \newcite{xu2018end} use a pointer network with a Seq2Seq architecture to handle unseen slot values. \newcite{Lee2019Jul} use a pre-trained BERT model~\cite{Devlin2018Oct} to encode slots and utterances and uses multi-head attention~\cite{vaswani2017attention} to find relevant information in the dialogue context for predicting slot values. \newcite{Wu2019May} introduce an encoder-decoder architecture with a copy mechanism, sharing all model parameters between all domains. ~\newcite{zhou2019multi} formulate multi-domain DST as a question answering task and use reading comprehension techniques to generate the answers by either span or value prediction.

\newcite{johnson2017google} propose single encoder-decoder models for zero-shot machine translation by encoding language and input sentence jointly, and \newcite{zhao2018zero} propose cross-domain zero-shot language generation using a cross-domain embedding space.

\begin{table*}[htb]
\centering
\small
\begin{tabular}{llll}
\toprule
{\bf From Abstract State}   & {\bf Agent Dialogue Act}    & {\bf User Dialogue Act}           & {\bf To Abstract State} \\
\midrule
Start     &                            & Greet                & Greeting         \\
                   &                            & Ask by name          & Info request     \\
                   &                            & Ask with constraints & Search request   \\
\hline
Greet              & Greet                      & Ask by name          & Info request     \\
                   &                            & Ask with constraints & Search request   \\
\hline
Search request     & Ask to refine search       & Provide constraints  & Search request   \\
\cline{2-4}
                   & Ask question               & Answer question      & Search request   \\
\cline{2-4}
                   & Propose constraint         & Accept constraint    & Search request   \\
                   &                            & Add constraints      & Search request   \\
\cline{2-4}
                   & Propose entity             & Accept               & Complete request \\
                   &                            & Add constraints      & Search request   \\
                   &                            & Reject               & Search request   \\
                   &                            & Ask slot question    & Slot question    \\
                   &                            & Ask info question    & Info question    \\
\cline{2-4}
                   & Empty search, offer change & Change constraints   & Search request   \\
                   &                            & Insist               & Insist           \\
\hline
Info request       & Provide info, offer reservation  & Accept             & Accept        \\
                   &                            & Provide reservation info & Accept        \\
                   &                            & Ask info question        & Info question \\
\hline
Info question      & Answer, offer reservation  & Accept                   & Accept        \\
                   &                            & Provide reservation info & Accept        \\
                   &                            & Thanks                   & Close conversation       \\
\hline
Slot question      & Answer, offer reservation  & Accept                   & Accept         \\
                   &                            & Add constraint           & Search request \\
\hline
Insist             & Repeat empty search        & Apologize                & Close conversation        \\
                   &                            & Change constraints       & Search request \\
\hline
Complete request   & Offer reservation          & Accept                   & Accept         \\
                   &                            & Thanks                   &   Close conversation         \\
\hline
Accept             & Ask missing slots          & Answer question                               & Complete transaction \\
\hline
Complete transaction & Execute                  & Ask transaction info     & Transaction info question \\
                     &                          & Thanks                   & Close conversation       \\
\cline{2-4}
                     & Error                    & Thanks                   & Close conversation       \\
\hline
Transaction info question & Answer        & Thanks                   & Close conversation       \\
\hline
Close conversation            & Anything else              & Thanks                   & End  \\
\bottomrule
\end{tabular}
\caption{Our abstract dialogue model for transaction dialogues. Each row represents one transition between abstract dialogue states.}
\vspace{-0.4cm}
\label{table:state-transition}
\end{table*}

\paragraph{Modelling of Dialogues.}
Previous work already  proposed general models of dialogues as finite state machines~\cite{jurafsky1997switchboard, bunt2017revisiting, yu2019midas}. Existing models are optimized for analyzing existing human conversations. Our dialogue model is the first suitable for synthesis, carrying enough information to continue the dialogue.

\newcite{DBLP:journals/corr/abs-1810-07942} previously proposed a different annotation scheme for dialogues, using a hierarchical representation scheme, instead of the more typical intent and slot. Their work is complementary to ours: our method of dialogue synthesis is applicable to any annotation scheme. In this paper, we focus on the existing annotation scheme used by the MultiWOZ dataset.

\section{Dialogue-Model Based Synthesis}
\label{sec:algo}
In this section, we first define abstract dialogue models, then describe how we can generate dialogues based on the model. We also describe the techniques we use to adapt training dialogues from other domains to the new domain.  

\subsection{Abstract Dialogue Model}
We define a {\em dialogue model} with finite sets of {\em abstract states}, {\em agent dialogue acts}, {\em user dialogue acts}, and {\em transitions}, defined below. The abstract dialogue for transactions we use in this paper is shown in Table~\ref{table:state-transition}. 

The {\em abstract states} capture the typical flow of a conversation in that model, regardless of the domain.  For example, a transaction dialogue model has states \Greet, \SearchRequest, \CompleteRequest, \CompleteTransaction, and \CloseConversation, etc.  Each domain has a set of {\em slots}; each slot can be assigned a {\em value} of the right type, a special \textsc{dontcare} marker indicating that the user has no preference, or a special ``\textsc{?}'' marker indicating the user is requesting information about that slot. Thus, we can summarize the content discussed up to any point of a conversation with a {\em concrete state}, consisting of an abstract state, and all the slot-value pairs mentioned up to that point.
Where it is not ambiguous, we refer to the concrete state as the {\em state} for simplicity.

All possible agent utterances in a dialogue model are classified into a finite set of {\em agent dialogue acts}, and similarly, all the possible user utterances into a finite set of {\em user dialogue acts}. Examples of the former are \GreetUser, \AskQuestion, \Answer, \OfferReservation; examples of the latter are \AskByName, \AddConstraints, \Accept, \Reject.

Each {\em transition} in the model describes an allowed {\em turn} in a dialogue. A transition consists of an abstract start state, an agent dialogue act, a user dialogue act, and an abstract end state. 

\subsection{Dialogues from an Abstract Model}
A {\em dialogue} is a sequence of turns, each of which consists of a start state, an agent utterance, a user utterance, and an end state.  We say that a dialogue belongs to a model, if and only if, 
\begin{enumerate}
\item 
for every turn, the start state's abstract state, the dialogue act of the agent utterance, the dialogue act of the user utterance, and the end state's abstract state constitute an allowed transition in the model.
\item
the slot-value pairs of each end state are derived by applying the semantics of the agent and user utterances to the start state.
\item
the first turn starts with the special \StartState state, and every turn's end state is the start state of the next turn, except for the last turn, where the end state is the special \EndState state.
\end{enumerate}

\subsection{Synthesizing a Turn with Templates}
We use templates to synthesize dialogues in a domain from an abstract dialogue model and a domain ontology. In this paper, we introduce {\em dialogue model templates} which specify with grammar rules how to generate a turn of a dialogue from a transition in the abstract model. They create possible agent and user utterances matching the agent and user dialogue acts in the transition, and they include a {\em semantic function} to ensure the utterances make sense given the input state.  For example, the user should ask about a slot only if its value is not already known. The semantic function returns an output state that matches the semantics of the utterances. The slot values of the output state are used as the annotation for the turn when training the dialogue state tracker.

As an example, the \SlotQuestion template shown in Fig.~\ref{fig:template-example} corresponds to the 13th transition in the dialogue model in Table~\ref{table:state-transition}. The following agent and user utterances, separated by a delimiting token \textless{}sep\textgreater{}, are examples of dialogue acts \ProposeEntity and \AskSlotQuestion.  They transition the abstract state \SearchRequest to the abstract state \SlotQuestion.  

{
\small
\begin{tabbing}
\=1234567\=\kill
\>State:\>$\textsc{SearchRequest}\ \text{restaurant}(\ldots)$\\
\>Agent:\>How about Curry Garden? It is an Indian \\
\>\>restaurant in the south of town. \textless{}sep\textgreater{}\\
\>User: \>Is it expensive?\\
\>State:\>$\textsc{SlotQuestion}\ \text{restaurant}(\ldots, \text{price}=\text{``?''})$
\end{tabbing}
}
In this case, the non-terminals \textsc{name}, \textsc{np}, \textsc{adj\_slot} are expanded into domain-dependent phrases
``Curry Garden'', ``Indian restaurant in the south of town'', and ``expensive'', respectively, and the results of their semantic functions, \textit{name, np, adj\_slot}, are (sets of) slot-value pairs: name = ``Curry Garden''; \{ food = ``Indian'', area= ``south'' \}; price = ``expensive''.  
The semantic function of \SlotQuestion checks that the input state does not already include a value for the price slot, and the price is not mentioned by the agent at this turn. It returns, as the new state, the old state with a ``?'' on the price. 

All the non-dialogue specific templates are introduced by \newcite{xu2020schema2qa}. We have extended this template library, originally intended for database queries, to return slot-value pairs as semantic function results.  Readers are referred to \newcite{xu2020schema2qa} for details. This library has four kinds of domain templates. \textit{Domain Subject Templates} describe different noun phrases for identifying the domain.  \textit{Slot Name Templates} describe ways to refer to a slot name without a value, such as ``cuisine'', ``number of people'' or ``arrival time''.  \textit{Slot Value Templates} describe phrases that refer to a slot and its value; they can be a noun phrase (``restaurants with Italian food''), passive verb phrase (``restaurants called Alimentum''),
  active verb phrase (``restaurants that serve Italian food''), adjective-phrase (``Italian restaurants''), preposition clauses (``reservations for 3 people''). Finally,  \textit{Information Utterance Templates} describe full sentences providing information,
  such as ``I need free parking'', or ``I want to arrive in London at 17:00''. These are domain-specific because they use a domain-specific construction (``free parking'') or verb (``arrive'').
  
Developers using our methodology are expected to provide domain templates, by deriving them manually from observations of a small number of in-domain human conversations, such as those used for the validation set.

\begin{figure}
\small
\begin{tabbing}
1234\=1234567890120\=12\=12\=\kill
\textsc{SlotQuestion} $:=$ ``How about'' \textsc{name} ``? It is a '' \textsc{np} ``.''\\
\>\>``\textless{}sep\textgreater{} Is it'' \textsc{adj\_slot} ``?'': \\
\>\>$\lambda(\textit{state}, \textit{name}, \textit{np}, \textit{adj\_slot}) \rightarrow \{$\\
\>\>\>\textbf{if} $\textit{adj\_slot} \in (\textit{state}.\text{slots} \cup \textit{np})$\\
\>\>\>\>\textbf{return }$\perp$\\
\>\>\>$\textit{state}.\text{abstract} = \textsc{SlotQuestion}$\\
\>\>\>$\textit{state}.\text{slots}[\textit{adj\_slot}.\text{name}] = \text{``?''}$\\
\>\>\>\textbf{return }$\textit{state}$\\
\>\>$\}$\\
\textsc{np} $:=$ \textsc{adj\_slot} \textsc{np} $: \lambda(\textit{adj\_slot}, \textit{np}) \rightarrow \textit{np} \cup \{ \textit{adj\_slot} \} $\\
\textsc{np} $:=$ \textsc{np} \textsc{prep\_slot} $: \lambda(\textit{np}, \textit{prep\_slot}) \rightarrow \textit{np} \cup \{ \textit{prep\_slot} \} $ \\
\textsc{np} $:=$ ``restaurant'' $: \lambda() \rightarrow \emptyset$\\
\\
\textsc{adj\_slot} $:=$ $\textsc{food}~~\vert~~\textsc{price} : \lambda(x) \rightarrow x$\\
\textsc{prep\_slot} $:=$ ``in the'' \textsc{area} ``of town'' $: \lambda(x) \rightarrow x$\\
\textsc{name} $:=$ $\text{``Curry Garden''}~~\vert~~\ldots : \lambda(x) \rightarrow \text{name} = x$\\
\textsc{food} $:=$ $\text{``Italian''}~~\vert~~\text{``Indian''}~~\vert~~\ldots : \lambda(x) \rightarrow \text{food} = x$\\
\textsc{area} $:=$ $\text{``north''}~~\vert~~\text{``south''}~~\vert~~\ldots : \lambda(x) \rightarrow \text{area} = x$\\
\textsc{price} $:=$ $\text{``cheap''}~~\vert~~\text{``expensive''}~~\vert~~\ldots : \lambda(x) \rightarrow \text{price} = x$\\
\end{tabbing}
\vspace{-0.4cm}
\caption{The \SlotQuestion template and other non-dialogue specific templates used to generate the example interaction.}
\vspace{-0.2cm}
\label{fig:template-example}
\end{figure}

\subsection{Synthesizing a Dialogue}

As there is an exponential number of possible dialogues, we generate dialogues with a randomized search algorithm. We sample all possible transitions uniformly to maximize variety and coverage. 
Our iterative algorithm maintains a fixed-size working set of incomplete dialogues and their current states, starting with the empty dialogue in the \StartState state. At each turn, it computes a random sample of all possible transitions out of the abstract states in the working set.
A fixed number of transitions are then chosen, their templates expanded and semantic functions invoked to produce the new concrete states. Extended dialogues become the working set for the next iteration; unextended ones are added to the set of generated results. The algorithm proceeds for a maximum number of turns or until the working set is empty. 

The algorithm produces full well-formed dialogues, together with their annotations. The annotated dialogues can be used to train any standard dialogue state tracker.


\subsection{Training Data Adaptations}
\label{sec:domain-adaptation}
We also synthesize new training data by adapting dialogues from domains with similar slots.
 For example, both restaurants and hotels have locations, so we can adapt a sentence like ``find me a restaurant in the city center'' to ``find me a hotel in the city center''.   
We substitute a matching domain noun phrase with the one for the new domain, and its slot values to those from the target ontology.  

We also generate new multi-domain dialogues from existing ones. We use heuristics to identify the point where the domain switches and we concatenate single-domain portions to form a multi-domain dialogue.

\section{Experimental Setting}
\label{sec:model}

\subsection{The MultiWOZ Dataset}
The MultiWOZ dataset~\cite{budzianowski2018large, eric2019multiwoz} 
is a multi-domain fully-labeled corpus of human-human written conversations. Its ontology has 35 slots in total from 7 domains. 
Each dialogue consists of a goal, multiple user and agent utterances, and annotations in terms of slot values at every turn. The dataset is created through crowdsourcing and has 3,406 single-domain and 7,032 multi-domain dialogues.

Of the 7 domains, only 5 have correct annotations and any data in the validation or test sets.
Following~\newcite{Wu2019May} we only focus on these 5 domains in this paper. The characteristics of the domains are shown in Table~\ref{table:dataset-breakdown}.
\begin{table*}
\small
\centering
\begin{tabular}{lrrrrr}
\toprule
                  & \bf Attraction & \bf Hotel & \bf Restaurant & \bf Taxi & \bf Train \\
\midrule
\# user slots   & 3              & 10        & 7              & 4        & 6         \\
\# agent slots  & 5              & 4         & 4              & 2        & 2         \\
\# slot values  & 167            & 143       & 374            & 766      & 350       \\
\hline
\hline
\# real dialogues  & 3,469        & 4,196     & 4,836          & 1,919    & 3,903     \\
\# in-domain turns & 10,549         & 18,330    & 18,801         & 5,962    & 16,081    \\
\# in-domain tokens & 312,569        & 572,955   & 547,605        & 179,874  & 451,521   \\
\hline
\hline
\# domain subject templates & 3  & 5         & 4              & 2        & 4         \\
\# slot name templates      & 15 & 17        & 21             & 18       & 16        \\
\# slot value templates           & 7  & 30        & 30             & 37       & 42        \\
\# information utterance templates      & 1  & 14        & 13             & 13       & 27        \\
\hline
\hline
\# synthesized dialogues & 6,636   & 13,300   & 9,901          & 6,771    & 14,092    \\
\# synthesized turns     & 30,274  & 62,950   & 46,062         & 35,745   & 60,236    \\
\# synthesized tokens    & 548,822 & 1,311,789 & 965,219       & 864,204  & 1,405,201 \\
\hline
\hline
transfer domain   & Restaurant & Restaurant & Hotel & Train & Taxi \\
overlapping slots & 2          & 6          & 6     & 4     & 4    \\
\bottomrule
\end{tabular}
\caption{Characteristics of the MultiWOZ ontology, the MultiWOZ dataset, the template library, and the synthesized datasets for the zero-shot experiment on the 5 MultiWOZ domains.
``user slots'' refers to the slots the user can provide and the model must track, while ``agent slots'' refer to slots that the user requests from the agent (such as the phone number or the address). Note that total number of dialogues is smaller than the sum of  dialogues in each domain due to multi-domain dialogues.}
\label{table:dataset-breakdown}
\end{table*}

\subsection{Machine Learning Models}
We evaluate our data synthesis technique on two state-of-the-art models for the MultiWOZ dialogue state tracking task, TRADE~\cite{Wu2019May} and SUMBT~\cite{Lee2019Jul}.
Here we give a brief overview of each model; further details are provided in the respective papers.

\paragraph{TRADE}
TRAnsferable Dialogue statE generator (TRADE) uses a soft copy mechanism to either copy slot-values from utterance pairs or generate them using an Recurrent Neural Network (RNN)~\cite{Sutskever2014Sep} decoder. This model can produce slot-values not encountered during training. The model is comprised of three main parts: an RNN utterance encoder which generates a context vector based on the previous turns of the dialogue; a slot-gate predictor indicating which (domain, slot) pairs need to be tracked, and a state generator that produces the final word distribution at each decoder time-step.

\paragraph{SUMBT}
Slot-Utterance Matching Belief Tracker (SUMBT) uses an attention mechanism over user-agent utterances at each turn to extract the slot-value information. It deploys a distance-based non-parametric classifier to generate the probability distribution of a slot-value and minimizes the log-likelihood of these values for all slot-types and dialogue turns.
Specifically, their model includes four main parts: the BERT~\cite{Devlin2018Oct} language model which encodes slot names, slot values, and utterance pairs, a multi-head attention module that computes an attention vector between slot and utterance representations, a RNN state tracking module, and a discriminative classifier which computes the probability of each slot value. The use of similarity to find relevant slot values makes the model depend on the ontology. Thus the model is unable to track unknown slot values.

\subsection{Software and Hyperparameters}

We used the Genie tool~\cite{geniepldi19} to synthesize our datasets.
We incorporated our dialogue model and template library into a new version of the tool. The exact version of the tool used for the experiment, as well as the generated datasets, are available on GitHub\footnote{\url{https://github.com/stanford-oval/zero-shot-multiwoz-acl2020}}. For each experiment, we tuned the Genie hyperparameters separately on the validation set.

For the models, we use the code that was released by the respective authors, with their recommend hyperparameters. For consistency, we use the same data preprocessing to train both TRADE and SUMBT.

\section{Experiments}
\label{sec:eval}
\subsection{Data synthesis}
Our abstract transaction dialogue model has 
13 abstract states, 15 agent dialogue acts, 17 user dialogue acts, and 34 transitions (Table~\ref{table:state-transition}).  We have created 91 dialogue templates for this model.
Dialogue templates were optimized using the validation data in the ``Restaurant'' domain. 

We also created domain templates for each domain in MultiWOZ. The number of templates and other characteristics of our synthesis are shown in Table~\ref{table:dataset-breakdown}. To simulate a zero-shot environment in which training data is not available, we derived the templates from only the validation data of that domain.  We did not look at in-domain training data to design the templates, nor did we look at any test data until the results reported here were obtained. In the table, we also include the domain we chose to perform domain adaptation (Section~\ref{sec:domain-adaptation}) and the number of slots from the adapted domain that are applicable to the new domain.

Note that the validation and test sets are the same datasets as the MultiWOZ 2.1 release.

\subsection{Evaluation On All Domains}
Our first experiment evaluates how our synthesized data affects the accuracy of TRADE and SUMBT on the full MultiWOZ 2.1 dataset.
As in previous work~\cite{Wu2019May}, we evaluate the \textit{Joint Accuracy} and the \textit{Slot Accuracy}. Joint Accuracy
measures the number of turns in which all slots are predicted correctly at once, whereas Slot Accuracy measures the accuracy of predicting each slot
individually, then averages across slots. Slot Accuracy is significantly higher than Joint Accuracy because, at any turn, most slots do not appear, hence predicting an empty slot yields high accuracy for each slot. Previous results were reported on the MultiWOZ 2.0 dataset, so we re-ran all models on MultiWOZ 2.1. 

Results are shown in Table~\ref{table:all-results}. We observe that our synthesis technique, which is derived from the MultiWOZ dataset, adds no value to this set.  We obtain almost identical slot accuracy, and our joint accuracy is within the usual margin of error compared to training with the original dataset. This is a sanity-check to make sure our augmentation method generates compatible data and training on it does not worsen the results. 

\begin{table}[t]
\small
\centering
\begin{tabular}{llrr}
\toprule
{\bf Model} & {\bf Synth.} & {\bf Joint} & {\bf Slot Acc.}\\
\midrule
\multirow{2}{*}{TRADE} & no  & 44.2 & 96.5 \\
                       & yes & 43.0 & 96.4 \\
\hline
\multirow{2}{*}{SUMBT} & no  & 46.7 & 96.7\\
                       & yes & 46.9 & 96.6\\
\bottomrule
\end{tabular}
\caption{Accuracy on the full MultiWOZ 2.1 dataset (test set), with and without synthesized data.}
\label{table:all-results}
\vspace{-0.5cm}
\end{table}
\begin{table*}[htb]
\small
\centering
\begin{tabular}{l|l|rr|rr|rr|rr|rr}
\toprule
&                                   & \multicolumn{2}{c|}{\bf Attraction} & \multicolumn{2}{c|}{\bf Hotel} & \multicolumn{2}{c|}{\bf Restaurant} & \multicolumn{2}{c|}{\bf Taxi} & \multicolumn{2}{c}{\bf Train}  \\
{\bf Model} & {\bf Training}        & {\bf Joint} & {\bf Slot}            & {\bf Joint} & {\bf Slot}       & {\bf Joint} & {\bf Slot}            & {\bf Joint} & {\bf Slot}      & {\bf Joint} & {\bf Slot}       \\
\midrule
\multirow{3}{*}{TRADE} 
& Full dataset
& 67.3 & 87.6
& 50.5 & 91.4
& 61.8 & 92.7
& 72.7 & 88.9
& 74.0 & 94.0 \\
\cline{2-12}
& Zero-shot
& 22.8 & 50.0
& 19.5 & 62.6
& 16.4 & 51.5
& 59.2 & 72.0
& 22.9 & 48.0 \\
& Zero-shot (Wu)
& 20.5 & 55.5 
& 13.7 & 65.6
& 13.4 & 54.5
& 60.2 & 73.5
& 21.0 & 48.9 \\
& Zero-shot (DM)
& \bf 34.9 & \bf 62.2
& \bf 28.3 & \bf 74.5         
& \bf 35.9 & \bf 75.6
& \bf 65.0 & \bf 79.9
& \bf 37.4 & \bf 74.5 \\
\cline{2-12}
& Ratio of DM over full (\%)
& 51.9 & 71.0 
& 56.0 & 81.5
& 58.1 & 81.6
& 89.4 & 89.9
& 50.5 & 79.3 \\
\hline
\hline
\multirow{4}{*}{SUMBT} 
& Full dataset 
& 71.1 &    89.1
& 51.8 &    92.2
& 64.2 &    93.1
& 68.2 &    86.0
& 77.0 &    95.0\\
\cline{2-12}
& Zero-shot
& 22.6    & 51.5       
& 19.8    & 63.3
& 16.5 & 52.1
& 59.5    & 74.9
& 22.5 & 49.2 \\
& Zero-shot (DM) 
& \bf 52.8 & \bf 78.9
& \bf 36.3 & \bf 83.7
& \bf 45.3 & \bf 82.8
& \bf 62.6 & \bf 79.4
& \bf 46.7 & \bf 84.2 \\
\cline{2-12}
& Ratio of DM over full (\%)
& 74.3 & 88.6 
& 70.1 & 90.8
& 70.6 & 88.9
& 91.8 & 92.3
& 60.6 & 88.6 \\
\bottomrule
\end{tabular}
\caption{Accuracy on the zero-shot MultiWOZ experiment (test set), with and without data augmentation. TRADE refers to~\newcite{Wu2019May}, SUMBT to~\newcite{Lee2019Jul}. ``Zero-shot'' results are trained by withholding in-domain data. ``Zero-shot (Wu)'' results are obtained with the unmodified TRADE zero-shot methodology, trained on MultiWOZ 2.1. ``Zero-shot (DM)'' refers to zero-shot learning using our Dialogue-Model based data synthesis. The last line of each model compares DM with full training, by calculating the \% of the accuracy of the former to the latter.}
\label{table:zero-shot}
\end{table*}

\subsection{Zero-Shot Transfer Learning}
\label{sec:zero-shot}

Before we evaluate zero-shot learning on new domains, we first measure the accuracy obtained for each domain when trained on the full dataset.  For each domain,
we consider only the subset of dialogues that include that particular domain and only consider the slots for that domain when calculating the accuracy. In other words, suppose we have a dialogue involving an attraction and a restaurant: a prediction that gets the attraction correct but not the restaurant will count as joint-accurate for the attraction domain.  This is why the joint accuracy of individual domains is uniformly higher than the joint accuracy of all the domains. 
Table~\ref{table:zero-shot} shows
that the joint accuracy for TRADE varies from domain to domain, from 50.5\% for ``Hotel'' to 74.0\% for ``Train''. The domain accuracy with the SUMBT model is better than that of TRADE by between 1\% and 4\% for all domains, except for ``Taxi'' where it drops by about 4.5\%.

In our zero-shot learning experiment, we withhold all dialogues that refer to the domain of interest from the training set, and then evaluate the joint and slot accuracies in the same way as before.  The joint accuracy with the TRADE model is poor throughout except for 59.2\% for ``Taxi''.  The rest of the domains have a joint accuracy ranging from 16.4\% for ``Restaurant'' to 22.9\% for ``Train''.  Upon closer examination, we found that simply predicting ``empty'' for all slots would yield the same joint accuracy. The zero-shot results for SUMBT are almost identical to that of TRADE. 

A different evaluation methodology is used by \newcite{Wu2019May} in their zero-shot experiment. The model for each domain is trained with the full dataset, except that all the slots involving the domain of interest are removed from the dialogue state. The slots for the new domain are present in the validation and test data, however. The method they use, which we reproduce here\footnote{\newcite{Wu2019May} reported results on MultiWOZ 2.0, while we report MultiWOZ 2.1. The results on the two datasets are all within 3\% of each other.}, has consistently higher slot accuracy, but slightly worse joint accuracy than our baseline, by 
1.9\% to 5.8\%, except for ``Taxi'' which improves by 1\% to 60.2\%.  

To evaluate our proposed technique, we add our synthesized data for the domain of interest to the training data in the zero-shot experiment.   Besides synthesizing from templates, we also apply domain adaptation. The pairs of domain chosen for adaptation are shown in Table~\ref{table:dataset-breakdown}, together with the number of slot names that are common to both domains. ``Taxi'' uses a subset of the slot names as ``Train'' but with different values. ``Attraction'', ``Restaurant'' and ``Hotel'' share the ``name'' and ``area'' slot; ``Restaurant'' and ``Hotel'' also share the ``price range'', ``book day'', ``book time'' and ``book people'' slots.
For slots that are not shared, the model must learn both the slot names and slot values exclusively from synthesized data.

Our dialogue-model based zero-shot result, reported as 
 ``Zero-shot (DM)'' in Table~\ref{table:zero-shot},  shows that our synthesized data improves zero-shot accuracy on all domains. For TRADE, the joint accuracy
improves between 6\% on ``Taxi'' and 19\% on ``Restaurant'', whereas for SUMBT, joint accuracy improves between 3\% on ``Taxi'' and 30\% on ``Attraction''. With synthesis, SUMBT outperforms TRADE by a large margin.  Except for ``Taxi'' which has uncharacteristically high joint accuracy of 65\%, SUMBT outperforms TRADE from 8\% to 18\%. This suggests SUMBT can make better use of synthesized data. 

\begin{figure}[t]
\centering
\includegraphics[width=\linewidth]{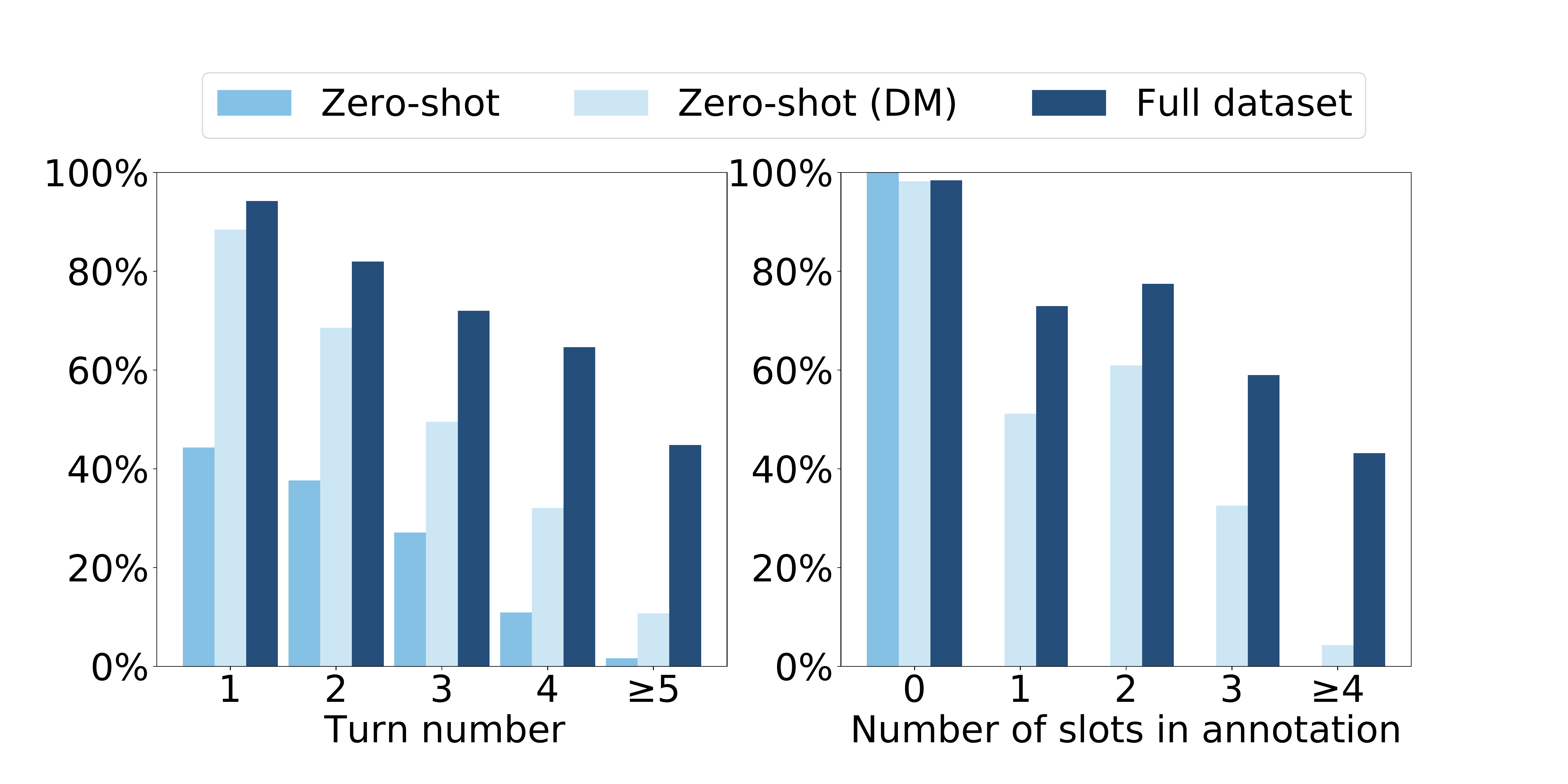}
\vspace{-0.5cm}
\caption{Breakdown of accuracy by turn number and number of slots of the TRADE model on the ``Restaurant'' domain. ``Zero-shot'' results are trained by withholding in-domain data, and ``Zero-shot (DM)'' is our data synthesis based on the Dialogue Model. ``Full dataset'' refers to training with all domains.}
\label{fig:error-analysis}
\end{figure}
\begin{figure*}[htb]
\centering
\includegraphics[width=\linewidth]{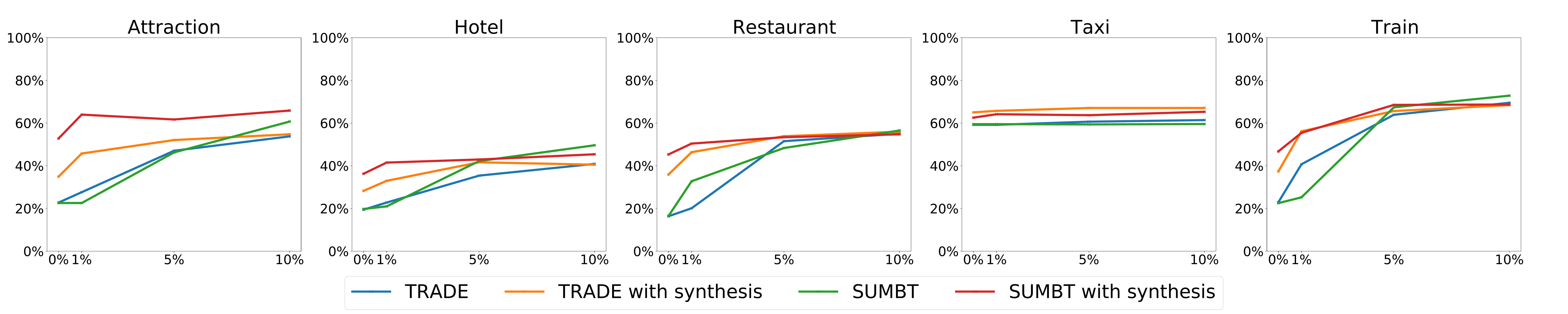}
\caption{Accuracy plots for the few-shot MultiWOZ experiments. X axis indicates the percentage of real target domain data included in training. Y axis indicates joint accuracy.}
\label{fig:few-shot}
\end{figure*}
To compare synthesized with real training data, 
we calculate how close the accuracy obtained with the synthetic data gets to full training. We divide the accuracy of the former with that of the latter, as shown in the last row for each model in Table~\ref{table:zero-shot}.  

Overall, training with synthesized data is about half as good as full training for TRADE, but is 2/3 as good as for SUMBT (the ratio is 61\% to 74\%, ignoring ``Taxi'' as an outlier).  This suggests that our synthesis algorithm is generating a reasonable variety in the dialogue flows; the pretrained BERT model, which imbues the model with general knowledge of the English language, is better at compensating for the lack of language variety in synthesized data.  Thus, the model only needs to learn the ontology and domain vocabulary from the synthesized data. Conversely, TRADE has no contextual pretraining and must learn the language from the limited dialogue data. This suggests that the combination of unsupervised pretraining and training on synthesized data can be effective to bootstrap new domains. 

\subsection{Error Analysis}
To analyze the errors, we break down the result according to the turn number and number of slots in the dialogues in the test set, as shown in Fig.~\ref{fig:error-analysis}.  We perform this analysis using the TRADE model on the ``Restaurant'' domain, which is the largest domain in MultiWOZ.  We observe that the baseline model achieves 100\% accuracy for turns with no slots, and 0\% accuracy otherwise.  The baseline results in the turn-number plot thus indicate the percentage of dialogues with all empty slots at each turn. 
It is possible for 5-turn dialogues to have all empty slots because a multi-domain dialogue may not have filled any slot in one domain. 

By and large, the accuracy degrades for both the ``full dataset'' model and the ``zero-shot (DM)'' model, with the latter losing more accuracy than the former when there are 3 or 4 slots.  
The accuracy drops almost linearly with increasing turn numbers for the full model.  This is expected because a turn is considered correct only if the full dialogue state is correct, and the state accumulates all slots mentioned up to that point. The results for the full and the zero-shot (DM) models look similar, but the zero-shot model has a larger drop in later turns.  Modeling the first few turns in the dialogue is easier, as the user is exclusively providing information, whereas in later turns more interactions are possible, some of which are not captured well by our dialogue model.

\subsection{Few-Shot Transfer Learning}

Following \newcite{Wu2019May}, we also evaluate the effect of mixing a small percentage of real training data in our augmented training sets.
We use a naive few-shot training strategy, where we directly add a portion of the original training data in the domain of interest to the training set.

Fig.~\ref{fig:few-shot} plots the joint accuracy achieved on the new domain with the addition of different percentages of real training data. The results for 0\% are the same as the zero-shot experiment. The advantage of the synthesized training data decreases as the percent of real data increases, because real data is more varied, informative, and more representative of the distribution in the test set.  The impact of synthesized data is more pronounced for SUMBT than TRADE for all domains even with 5\% real data, and it is significant for the ``Attraction'' domain with 10\% real data. This suggests that SUMBT needs more data to train, due to having more parameters, but can utilize additional synthesized data better to improve its training. 


\section{Conclusion}
\label{sec:conclusion}

We propose a method to synthesize dialogues for a new domain using an abstract dialogue model, combined with a small number of domain templates derived from observing a small dataset.
For transaction dialogues, our technique can bootstrap new domains with less than 100 templates per domain, which can be built in a few person-hours. With this little effort, it is already possible to achieve about 2/3 of the accuracy obtained with a large-scale human annotated dataset. Furthermore, this method is general and can be extended to dialogue state tracking beyond transactions, by building new dialogue models. 

We show improvements in joint accuracy in zero-shot and few-shot transfer learning for both the TRADE and BERT-based SUMBT models. Our technique using the SUMBT model improves the zero-shot state of the art by 21\% on average across the different domains. This suggests that pretraining complements the use of synthesized data to learn the domain, and can be a general technique to bootstrap new dialogue systems.

We have released our algorithm and dialogue model as part of the open-source Genie toolkit, which is available on GitHub\footnote{\url{https://github.com/stanford-oval/genie-toolkit}}.

\section*{Acknowledgments}
This work is supported in part by the National Science Foundation
under Grant No. 1900638; Mehrad Moradshahi is supported by a Stanford Graduate Fellowship. 

\bibliography{paper}
\bibliographystyle{acl_natbib}

\end{document}


\maketitle

\section{Hyperparameters}

\subsection{Genie}

When generating the first turn of the dialogues, we set:
\begin{itemize}
\item Pruning size: 50,000
\item Max depth: 9
\end{itemize}

When generating subsequent turns, we set:
\begin{itemize}
\item Pruning size: 1,000
\item Minibatch size (working set size): 10,000
\item Max depth: 6
\item Max turns: 6
\end{itemize}

For the zero-shot experiment, we sample 6\% of the synthesized dialogues. For the full experiment, we sample 3\%.

\subsection{TRADE}

All hyperparameters are as in the TRADE paper, except batch size is set to 8.

\subsection{SUMBT}

All hyperparameters are as recommended in the SUMBT code release.